\RequirePackage{fix-cm}
\documentclass[smallcondensed,final,natbib]{svjour3}       
\smartqed  
\usepackage{graphicx}
\usepackage{epstopdf}
\usepackage{color}
\usepackage{algorithm2e}
\usepackage{algorithmic}
\usepackage{amsmath}
\newcommand{\PSOC}{CriPS}
\begin{document}

\title{\PSOC:\\ Critical Dynamics in Particle Swarm Optimisation
}


\author{Adam Erskine         \and
        J.~Michael Herrmann 
}


\institute{A.~Erskine and J.~M.~Herrmann\at
              The University of Edinburgh, School of Informatics, Institute for Perception, Action and Behaviour, 10 Crichton St, Edinburgh EH8 9AB, U.K. \\
              \email{a.erskine@sms.ed.ac.uk, michael.herrmann@ed.ac.uk}           
}

\date{Received: date / Accepted: date}

\maketitle

\begin{abstract}
Particle Swarm Optimisation (PSO) makes use of a dynamical system for solving a search task. 
Instead of adding search biases in order to improve performance in certain problems, we aim to remove 
algorithm-induced scales by controlling the swarm with a mechanism that is scale-free except
possibly for a suppression of scales beyond the system size. In this way a very promising 
performance is achieved due to the balance of large-scale exploration and local search. 
The resulting algorithm shows evidence for self-organised criticality, 
brought about via the intrinsic dynamics of the swarm as it interacts with the objective 
function, rather than being explicitly specified. The Critical Particle Swarm (\PSOC) can 
be easily combined with many existing extensions such as chaotic exploration, additional 
force terms or non-trivial topologies.
\end{abstract}


\section{Introduction}
\label{introduction}
Particle swarm optimisation (PSO) is a metaheuristic method for obtaining solutions to optimisation problems~\citep{kennedy1995particle}. 
Inspired by the cooperative behaviours of flocks of birds or schools of fish, it employs 
a number of particles as a swarm of potential solutions. 
PSO argues that better solutions are located by balancing the exploitation of known good solutions with the exploration of areas of the problem space as yet under visited.
So, the dynamics of the particles are influenced by good solutions already found, but also includes overshoots and some randomness which
help to explore for yet better solutions. In order to perform optimally, the algorithm has to balance local search, i.e.~the 
exploitation of existing knowledge of the problem space, with exploration of potentially underexploited areas.
This balance is achieved via the parameterisation of the algorithm. 

Because metaheuristic algorithms are often applied to problems with little formal specification, trial-and-error search remains the only general 
option for parameter tuning. Furthermore, for the off-line identification of parameters 
typically only part of the solution space can be used,
often with a preference for quick progress in the solution space. 
This may not yield the best possible performance in near-optimal regions where 
a more fine-grained search may be required.

A number of mechanisms for parameter adaption in PSO has been proposed. Various controls on particle velocities have been used e.g. a simple maximum value~\citep{kennedy1998behavior} or a decreasing inertia 
weight~\citep{fan2007decreasing} in order to guide the algorithm from an
exploring to an exploiting behaviour. Since this may not prevent the stagnation at a
suboptimal solution, it has been proposed to explicitly increase the diversity, e.g.~by using
repulsion~\citep{riget2002diversity,chowdhury2013mixed}
or random velocities~\citep{garcia2009introducing}. It is further possible to vary the topology
of the swarm~\citep{kennedy2002population} but we will not explore this option here. Also the
field of hyperheuristic, memetic, and hybrid algorithms cannot be touched within the present scope, although
a comparison could be interesting.

A promising approach has been 
taken by~\citet{xie2002dissipative}
who introduced chaos (``negative entropy'') such that a
balance between the local and global search ability was reached. The algorithm was shown to perform 
well in the balanced regime but the right amount of quasi-random perturbation had to be 
identified by exhaustive search. See also the paper by \citet{liu2005improved} for a similar study.
We aim for a self-adaptive algorithm that solves the parameter
selection problem on-line using the properties of the problem it experiences. 
Self-adaptive algorithms have been proposed for differential evolution~\citep{qin2005self} and also
for PSO~\citep{wang2011self}. These algorithms use an adaptive probability model to choose
from a number of strategies in a success-dependent fashion.



Our aim is to modify the swarm dynamics of the PSO algorithm such that exploration and exploitation of the problem space are balanced automatically and on-line during the optimisation process. Intuitively, we induce a more stable behaviour should the swarm tend to diverge, but change the parameters of the algorithm towards the unstable regime if the swarm is likely to collapse. 
In this way we hope to achieve an optimal compromise between local fine-grained 
search near candidate optima and large scale exploration. 
The swarm will stay near a critical regime between stability and instability
which sometimes is compared to the \emph{edge of chaos}~\citep{langton1990computation}, although
the term \emph{criticality} seems more suitable here. We could say that
the compromise of local search and sudden exploratory excursions
is akin to the dynamics a sand pile:
As grains of sand are dropped onto a sandpile, the inclination of the slope approaches a critical value.
The addition of a single grain may release an avalanche, frequently these will be small slides, 
but occasional giant slips will occur. For an ideal sandpile, the avalanches will not show a typical size such that 
the distribution of the number of grains in an avalanche is scale-free and follows a power law within the region from 
single grains up to the size of the pile. Although the model may not describe very well real sandpiles, it has 
illustrated the interesting properties of criticality and lead to studies in various fields~\citep{bak1997nature}.

In particular, in the context of neural dynamics
the  functional benefits of critical dynamics for information processing in the brain
have been discussed, see the overview by~\citet{Shew2013functional}. 
These studies gave reason to the present study on the question whether 
criticality can be beneficial as well in dynamical systems for optimisation and search.

Another motivation of our approach can be drawn from Bayesian theory. 
An algorithm that traverses the search space in a step-wise fashion, specifies implicitly or explictly a prior for this step size,
i.e.~an assumption that steps should have a certain length distribution 
in order to optimally approach the goal of the search. If it is known that the goal is within a certain box, the optimal distribution
will assign zero probability to lengths larger than the size of the box. If the objective function is smooth, then often a minimal step 
size can be fixed which also may help to speed up the search. If, however, no information on the problem scale is available, then the 
best choice to use a non-informative prior that allows for step sizes on all scales. The shape of such a prior is given by a power-law
with an exponent of unity, but also for other exponents the step lengths can still be scale-free. The exponent may then be determined by
additional information on dimensionality of the problem or on the general structure of the objective function: 
If the objective has a number of smooth local optima, then relatively more of the 
larger jumps are needed than for a function that is shaped like a fractal. 
A prior does not necessarily show up in the results which are a combination of the properties of the specific objective function and 
any prior assumption that entered the algorithm. We will study this in the results section.

While criticality in piles of granular matter happens around specific angles, 
criticality of a swarm of particles can be described in a
much easier way: If the swarm does neither collapse nor diverge it is critical. We will be specific below in section~\ref{sub:crit} after we have described the standard PSO algorithm~\ref{sub:PSO}. 
We will then introduce our algorithm which combines PSO and criticality (\ref{sect:comb}) and 
analyse its performance in various respects in Sect.~\ref{sect:res}. The discussion (Sect.~\ref{sect:disc}) 
addresses the relation to the theory of criticality and to other optimisation algorithms that are based on criticality.




\section{Background}


\subsection{Particle swarm optimisation}\label{sub:PSO}

In PSO, each particle represents a solution to a given problem. Its state, i.e.~position and velocity,
is continuously changed in order to arrive at better states or to track a dynamic solution. The state is updated using the following 
rules~\citep{kennedy1995particle}
\begin{eqnarray}
\mathbf{v}_{i}(t+1)&=&\omega\mathbf{v}_{i}(t)+\alpha_{1}\mathbf{R}_{1}(\mathbf{p}_{i}-\mathbf{x}_{i})+\alpha_{2} \mathbf{R}_{2}(\mathbf{g}-\mathbf{x}_{i})
\label{eq:PSOVelocityUpdate}\\
\mathbf{x}_{i}(t+1)&=&\mathbf{x}_{i}(t)+\mathbf{v}_{i}(t+1)
\label{eq:PSOPositionUpdate}
\end{eqnarray}
The velocity update is a linear combination of three contributions: An inertial term parameterised by $\omega$, 
a pull towards the personal best location $\mathbf{p}_{i}$ parameterised by $\alpha_{1}$, 
and a pull towards the global best location $\mathbf{g}$ parameterised by $\alpha_{2}$. 
The symbols $\mathbf{R}_{1}$ and $\mathbf{R}_{2}$ denote diagonal matrices whose non-zero entries are uniformly distributed in the unit 
interval.\footnote{This implements a component-wise multiplication of the difference vectors with a random vector and is 
known to introduce a bias into the algorithm as particles prefer to stay near the axes of the 
problem space~\citep{janson2007trajectories,spears2010biases}.
We have retained this form of stochasticity in our algorithm for the purpose of comparability.}
There are PSO variants that explore the effect of changing what constitutes the global best term. In some systems local best values are used instead. Here sub-swarms communicate the best values they have seen. The global best version is therefore the fully connected version of this. A review of PSO variants \citep{bratton2007defining} concludes that a local connectivity of two outperforms the global best approach at least in the long term. 
One may also consider the global best as the best value obtained in the previous iteration rather than in the whole run of the algorithm. In this case, long term memory is provided by the particles' personal best values only. 


Other improvements to PSO often concentrate on mechanisms to improve the selection of the parameters $\omega$, $\alpha_{1}$, $\alpha_{2}$ 
(and others in variants of the algorithm).
Updates are often made as the algorithm progresses. Starting with large values (particularly of the $\omega$ term)
ensures that the particles are initially encouraged to explore the problem space. 
Later the parameter value is decreased ensuring that the particles gradually shift to exploiting the good locations they have already discovered. Numerous mechanisms to vary these parameters have been explored including constant, random, increasing, decreasing and chaotic~\citep{bansal2011inertia}. 
Typically, variants are tested against a number of test functions which have known optima. 
A number of criteria for success are measured (including average error, average number of iterations of obtain an optimum solution and minimum error). No one strategy succeeds against all criteria. 

Numerous further variations to the basic algorithm have been suggested, 
often inspired by the behaviour of other animals such as bat \citep{yang2010new}, 
cuckoo \citep{yang2009cuckoo} and fish \citep{wang2005improved} using respectively, 
echolocation, brood parasitism or leaping motion as inspirations. 
We note in particular the cuckoo search algorithm for its use 
of L\'{e}vy flights which enable the swarm to perform power-law distributed local searches. 
The distribution of the step lengths will be similar as in the algorithm presented here, however, 
in our algorithm this distribution is not injected but arises from the interaction between 
the particles' intrinsic dynamics and the objective function such that the dynamics can still adapt 
to the particularities of the problem.


Our approach is to make the algorithm responsive to its environment so that it is able to self-tune to the current problem during the learning phase. The technique presented here was first explored by~\citet{cordero2012}. The diversity of the particle swarm is monitored and used as a feedback signal to modify the swarm's dynamics. The swarm's response to the problem space shapes the swarm. A measure of the swarm is then used to change future behaviour.

\subsection{Criticality\label{sub:crit}}

Criticality in equilibrium thermodynamics is used to refer to the transition point between phases in the system. At the system transition temperature small perturbations to the state of the system can propagate throughout the whole system \citep{jensen1998self}. The propagation of magnetic spins in a ferromagnetic material at its critical temperature shows this. More widely the term may refer to any dynamical system which behaves in a manner like this \citep{bak1988self}. Such systems (unlike the thermodynamic equilibrium systems) will drive themselves into this state i.e. will self-organise. A characteristic of such systems is that certain properties of the system follow power-law distributions. Thus the frequency distribution of avalanche sizes in the much studied sand pile model \citep{bak1988self} shows the characteristic spread of a power-law distribution: few very large events; many small events.

This dynamics is seen in many natural and man-made systems. Examples include earthquake magnitudes \citep{Olami1992Earthquakes}, forest fire models \citep{Drossel1992ForestFire}, punctuated evolution \citep{bak1993punctuated}, neuronal avalanches \citep{beggs2003neuronal, shew2011information}. Many reports of systems that show critical behaviour focus on the presence of a power-law distribution of some property of the system. There is however more to criticality. Any power-law distributed property should present over several decades of scale, there should be multiple properties that show such a relationship and these should be mathematically inter-related. Additionally the system should be tunable through the critical point by some parameter of the system. The properties of such systems are discussed in an accessible manner by \citep{beggs2012being}. The presence of criticality in a system can be shown to optimise the system in some way. This \citep{shew2011information} showed the criticality of cortical neuronal avalanches results in optimal information capacity and transmission.

\subsection{Critical swarm}\label{sect:comb}

The Critical Particle Swarm (\PSOC) algorithm modifies the PSO parameter using feedback from the swarm dynamics. This results in a continuous exploration of the problem space whilst exploiting the knowledge that has been found already. 
Additionally by having the swarm self-organise itself into this active state the need to fine tune the parameters for a given problem would be avoided.

Criticality and PSO have been combined via mechanisms different to the one we are going to explore. 
A sandpile avalanche-like mechanism was employed \citep{lovbjerg2002extending}. Here, each particle was given a criticality parameter. This was a simple integer counter, incremented whenever particles were close together. Whenever this parameter rose above a threshold value two steps were taken. First the criticality value of the particle was redistributed to nearby particles (each gaining a single increment). If there were too few nearby particles the remaining value was lost from the system (akin to sand pouring off the each of the table). Second, the particle was randomly repositioned in the problem space. The redistribution of criticality values can obviously lead to further redistributions in an avalanche-like fashion. 
It is somehow elegant that feedback from the swarm's own behaviour is used to modify the future behaviour of the swarm. Improvements over standard PSO and some evidence that the swarm behaved in a critical manner were shown. However, a further parameter is introduced: a distance threshold for incrementing the criticality parameter. 

A Bak-Sneppen algorithm augmented PSO was described \citep{fernandes2012controlling}. Here a second algorithm is used to generate power-law distributed random numbers (similarly to the above mentioned cuckoo search algorithm) and adding them to the parameters in a statistically balanced way.
Additionally, local search about the particles position was driven by random numbers from the same distribution. 
This power-law distribution may be created offline adding little or no overhead to the algorithm. Favourable results are demonstrated in comparison to standard PSO. It has not been shown whether
the swarm is behaving in a self-organised critical manner, but  we have to note that the aim was to improve the PSO results and not necessarily to produce a swarm acting in a critical manner.

\section{Methods}

\subsection{The \PSOC{} algorithm}

In order to maintain the responsivity of the swarm to the objective function,
we will control the parameters based on a measure of the swarm's diversity.
There are many choices for such a measure we could use: The average interparticle distance, the average distance from each particle to the centroid of the swarm, or the average velocity norm of all the particles. All these metrics provide some indication of whether the swarm is expanding (and thus exploring) or contracting (and thus exploiting) within the problem space. 

We use the change in the metric value between iterations to provide a feedback signal to update the parameter values. For any metric $S$ as listed above, the change in its value is
\begin{equation}
\Delta S=S(t+1)-S(t).\label{eq:MetricChange}\end{equation}
Rather than applying this directly we squash it via a sigmoid function scaled to return [-1,1]. This ensures that small changes in swarm size have less impact on than large changes. 

We update the parameters of the PSO algorithm using 
\begin{equation}
\theta(t+1)=\theta(t)-\varepsilon \, f(\Delta S),
\label{eq:AbsoluteParamUpdate}
\end{equation}
where $\theta$ is any of the PSO parameters: $\omega$, $\alpha_1$, or $\alpha_2$. 
The parameter $\varepsilon<1$ controls the size of parameter updates. 
Finally, the rescaled sigmoid is given by
\begin{equation}
f(\Delta S) = \tanh\left(-\frac{\Delta S}{2 \sigma}\right)
\label{eq:sigmoid}
\end{equation}
where $\sigma$ is a scaling factor that is related to the extension of the search space in units of $\Delta S$.
\PSOC{} is shown in Algorithm~\ref{alg:PSOC}.

\vspace*{1\baselineskip}

\begin{algorithm}
\normalsize
\begin{enumerate}
\item Initialise PSO particles and parameters
\item Calculate $\text{S}$, the swarm metric
\item Calculate updated swarm velocities and positions using Eqs.~\eqref{eq:PSOVelocityUpdate} and \eqref{eq:PSOPositionUpdate}
\item Calculate swarm metric
\item Calculate $\Delta\text{S}$, the change in the swarm metric
\item Update parameters using Eqs.~\eqref{eq:MetricChange} and 
\eqref{eq:AbsoluteParamUpdate}
\item Evaluate whether goal reached
\item If goal reached or maximal time exceeded, exit; else go to step 3.
\end{enumerate}
\caption{\PSOC{} algorithm}\label{alg:PSOC}
\end{algorithm}

\subsection{Evaluation}

The results we present here were obtained using a Schwefel test function. As we wished remove any boundary conditions that might impede the motion of the particles we defined this to have a large (poor) fitness value outside the region of interest. Thus an $N$ dimensional Schwefel function returns a fitness value given by
\begin{equation}\label{SchwefelFunction}
f(x)=418.9829 N-\sum\limits_{i=1}^N x_i\sin(\sqrt{|x_i|}), \,\, 
x_i\in{[-500,500]}, \forall i 
\end{equation}
but outside this region the fitness values are simply defined as
\begin{equation}\label{SchwefelFunctionOut}
f(x)=500 N, \,\, x_i\notin{[-500,500]}.
\end{equation}
This ensures that the PSO algorithm is free to explore anywhere within the $N$ dimensional space, but should only find prospective optima within the region of interest. Particles are randomly initialised only within this region. 

Our initial setup was to use the absolute parameter update rule \eqref{eq:AbsoluteParamUpdate}, the velocity norm metric and initial parameter values of $\omega = 0.815$, $\alpha_1 = 1.0$, $\alpha_2 = 1.0$, and $\varepsilon = 0.15$. 

We set the parameter $\sigma$ in Eq.~\ref{eq:sigmoid} to a fifth of the maximum distance in any one dimension of the problem space. 
For the domain of definition of the Schwefel function ($[-500,500]^N$) this is 200. This value was determined via experiment. 

We further tested the algorithm against the Griewank function and a variant of the Griewank function where we shifted the optimum position to avoid it sitting at the origin. These functions were also defined with a large fitness value outside their regions of interest. These were used to check that we had not just created an ideal Schwefel function solver. The algorithms behaviour remained the same for these new functions. This result was obtained with no alteration to any parameter of the algorithm. Clearly, more extensive evaluation of the algorithm is necessary and will be provided in a forthcoming paper.

\section{Results}\label{sect:res}
Our aim was to modify the swarm dynamic of the PSO algorithm 
in order 
to overcome the problems of stagnation and divergence of the swarm. To this end we need to show that our algorithm 
continues to locate better solutions so long as the algorithm is running or until an optimal solution is found.
It would be beneficial for the algorithm to obtain reasonably good results quickly as well, but this is not the chief design aim.
Nevertheless, the swarm should show that exploration can extend over the full region of interest within the problem space. 
Finally, we want 
to show that the quasi-random changes of the size of the swarm follow a critical distribution as this implies a
search across all length scales. 

\subsection{Comparison with PSO}
\begin{figure}[htb]
\centering
\includegraphics[scale=0.5]{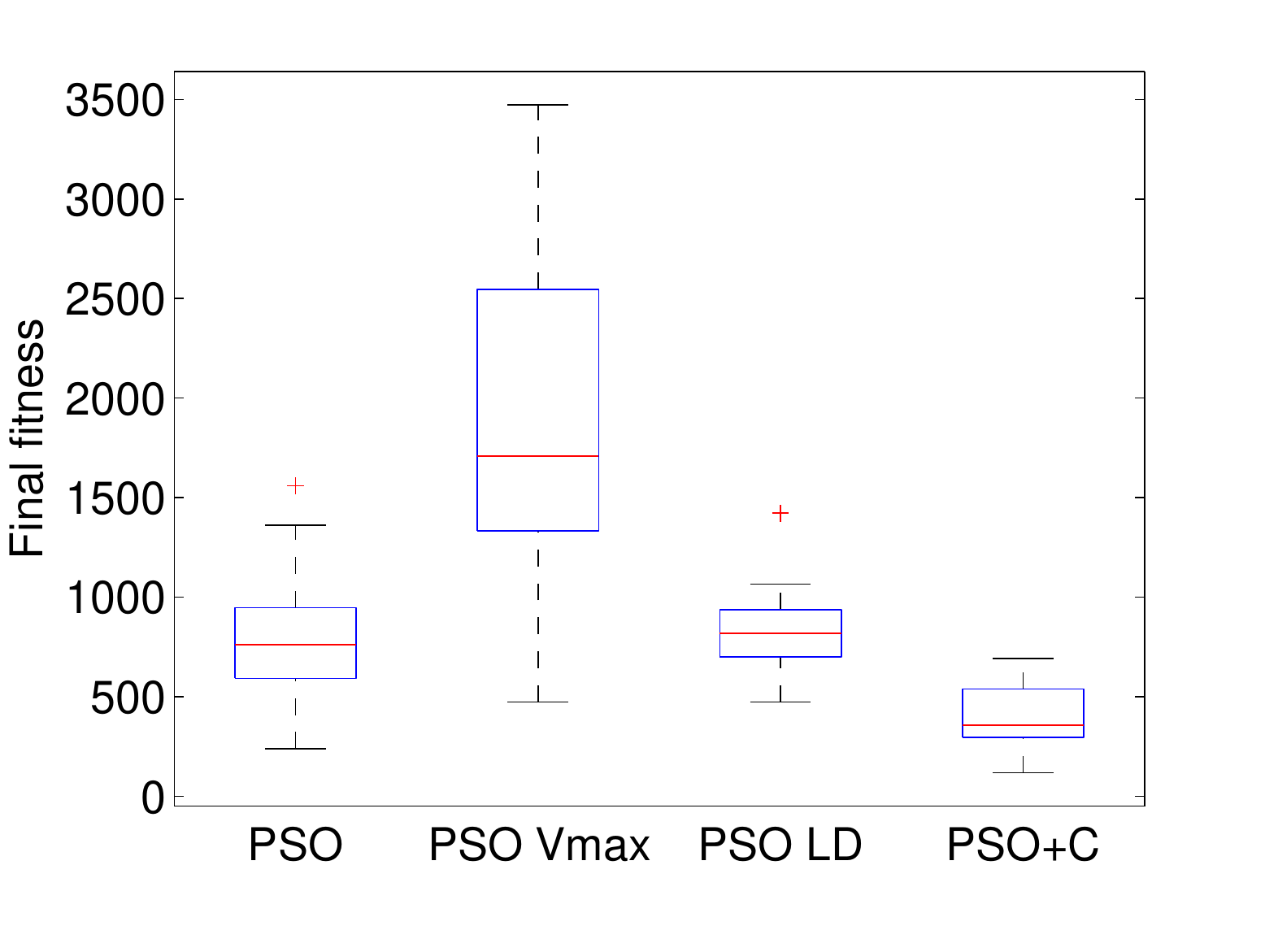}
\caption{Comparison of the algorithms. 
Objective function is 20 dimensional Schwefel function. 
Each algorithm uses 25 particles. 
Each run was 50,000 iterations long and was repeated 20 times. 
From left to right we have results for our standard PSO, PSO with a maximum particle velocity imposed (PSO Vmax), 
PSO with linear descent (PSO LD) of the $\omega$ parameter from 07 to 0.4, and the \PSOC{} algorithm.}\label{fig:AlgorithmComparison}
\end{figure}

To test the performance of the algorithm we use the Schwefel function defined in Eqs.~\eqref{SchwefelFunction} and \eqref{SchwefelFunctionOut}. This has an optima at location 420.9687 in each dimension. 
We choose this function as some formulations of PSO show a bias toward optima at the origin due to which
certain test functions may become trivial. 
We use an $N$ dimensional search space with $N = 20$ to provide a reasonably challenging problem. 
The term $-418.9829 N$ term in Eq.~\eqref{SchwefelFunction} ensures that the optimum of the function has the value zero at any number of dimensions. For the comparison, all runs will contain swarms with 25 particles. 
Initial comparisons are made between our algorithm and a PSO algorithm using fixed parameter values. The parameter values have been obtained using trial and error, better solutions may be possible. 
Our PSO formulation implements no maximum velocity to match \PSOC's absence of limits, but we also ran a version of PSO with a maximum velocity of 50 units. We make similar comparisons with a linear descent PSO using parameter values mentioned by \citet{lovbjerg2002extending}, i.e.~$\omega_{start} = 0.7, \omega_{end} = 0.4, \alpha_{1} = \alpha_{2} = 2$. 
Again better parameter values may be obtainable for this test function. We run the algorithm for 50000 iterations, twenty times. 
We see in Fig.~\ref{fig:AlgorithmComparison} that \PSOC{} outperforms the comparison algorithms with our test conditions.
Similar results were obtained for the \PSOC{} algorithm started with different initial parameters ($\omega$=0.7, $\alpha_1$ = $\alpha_2$ = 2).

\begin{figure}[htb]
\centering
\includegraphics[scale=0.5]{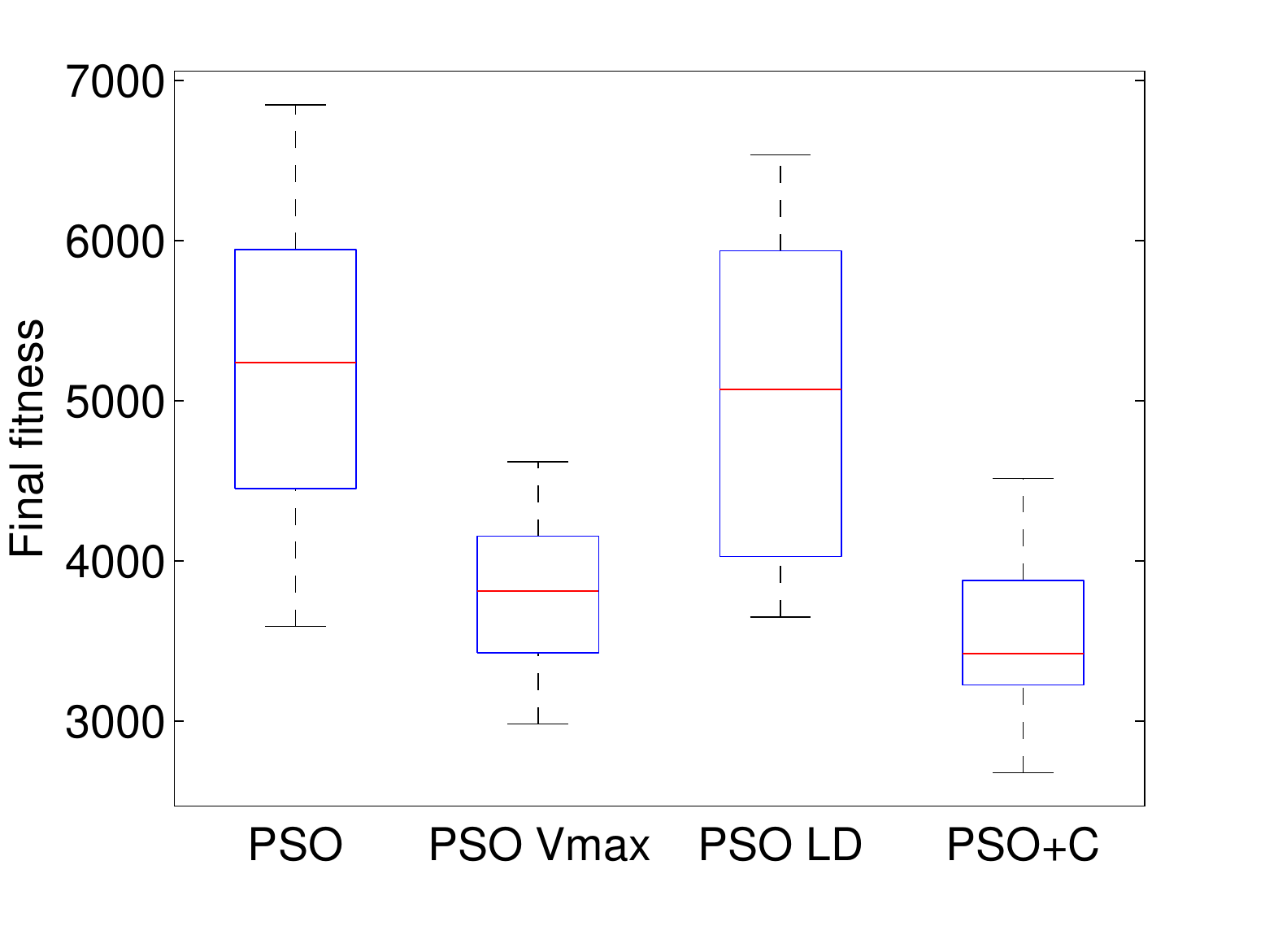}
\caption{Comparison of the algorithms. 
Objective function is 20 dimensional Schwefel function. 
Each algorithm uses 25 particles. 
Each run was 1,000 iterations long and was repeated 20 times. 
From left to right we have results for our standard PSO, PSO with a maximum particle velocity imposed (PSO Vmax), 
PSO with linear descent (PSO LD) of the $\omega$ parameter from 07 to 0.4, and the \PSOC{} algorithm.}\label{fig:AlgorithmComparison1000}
\end{figure}

The early success of the algorithm is explored by comparing the fitness values found by the various algorithms after 1000 iterations. We see in Fig.~\ref{fig:AlgorithmComparison1000} the \PSOC{} algorithm compares favourably with the other algorithms after relatively few iterations.

\subsection{Local search}
\begin{figure}[h!]
\centering
\includegraphics[scale=0.5]{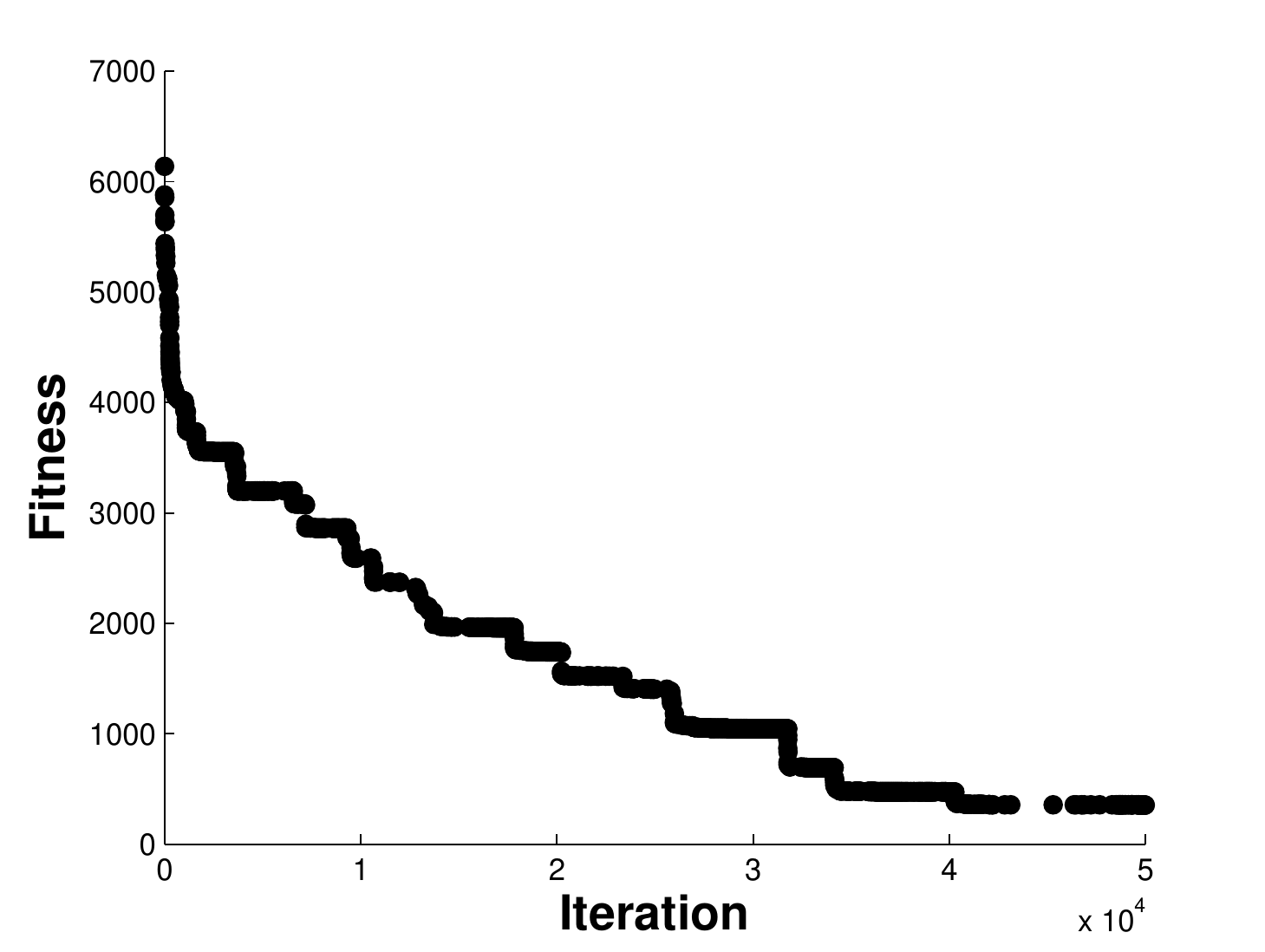}
\caption{\PSOC{} continuing improvements.  Each plotted point indicates a new global best position has been located. \PSOC{} continues to find new global best locations even as the number of iterations increases. Although small, improvements are still apparent at the end of the 50,000 iterations.}\label{fig:PSOC_ContinuingImprovements}
\end{figure}

\begin{figure}[h!]
\centering
\includegraphics[scale=0.5]{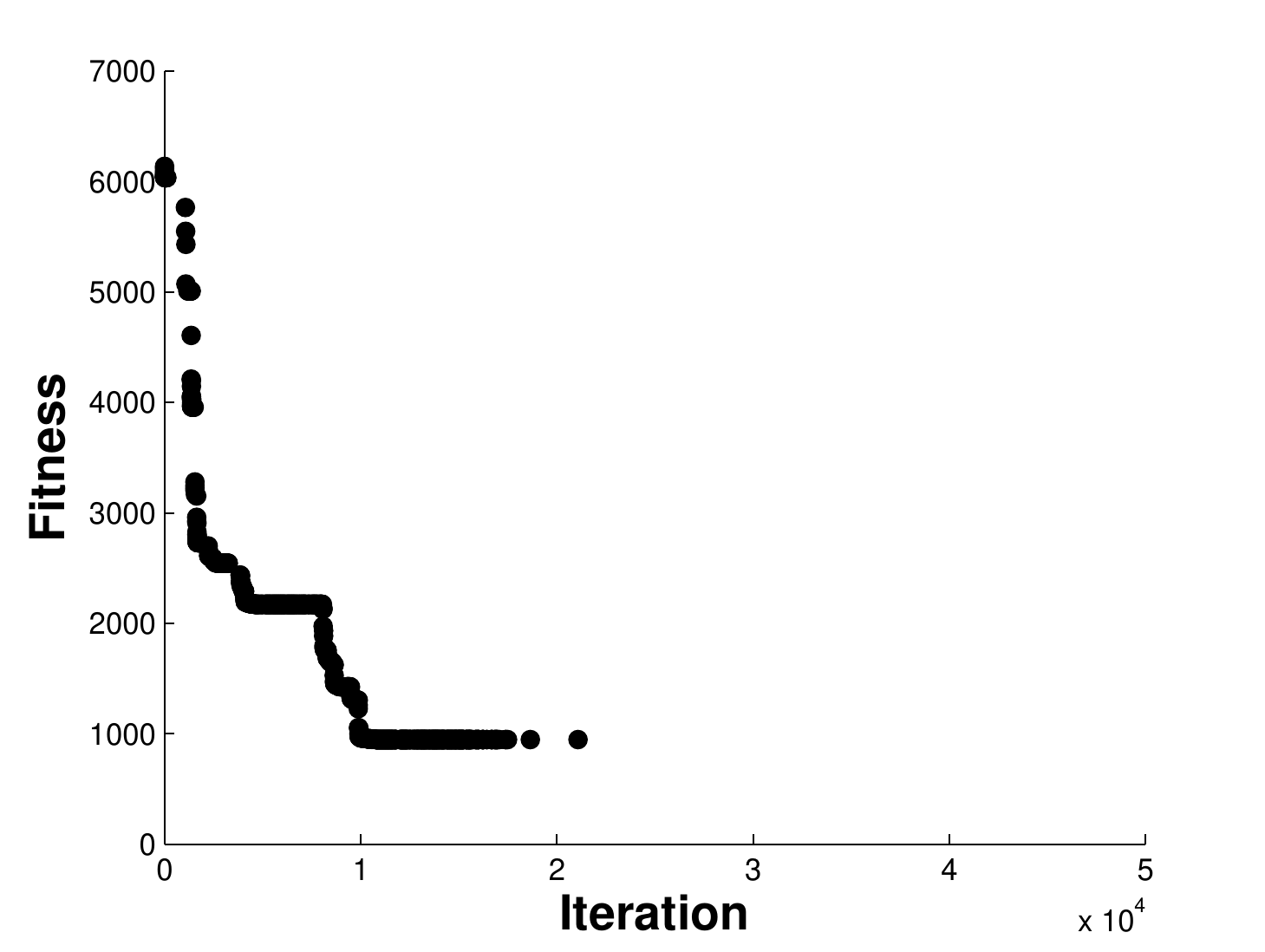}
\caption{PSO continuing improvements.  Each plotted point indicates a new global best position has been located. PSO finds these with lower frequency as the number of iterations increases. Improvements become less frequent and there are none in the last 28,000 or so iterations.}\label{fig:PSO_ContinuingImprovements}
\end{figure}

Each algorithm was run for 50000 iterations. Fig.~\ref{fig:PSOC_ContinuingImprovements} shows that \PSOC{} 
continues to locate better solutions throughout the entire run. 
The standard PSO (Fig.~\ref{fig:PSO_ContinuingImprovements}) shows that whilst better results are found frequently early on, after approximately 22000 iterations, in this run of the algorithm, no further improvements are found. Further executions of the algorithm show the same pattern.


\subsection{Comparison with random search}

\begin{figure}[h!]
	\centering
	\includegraphics[scale=0.5]{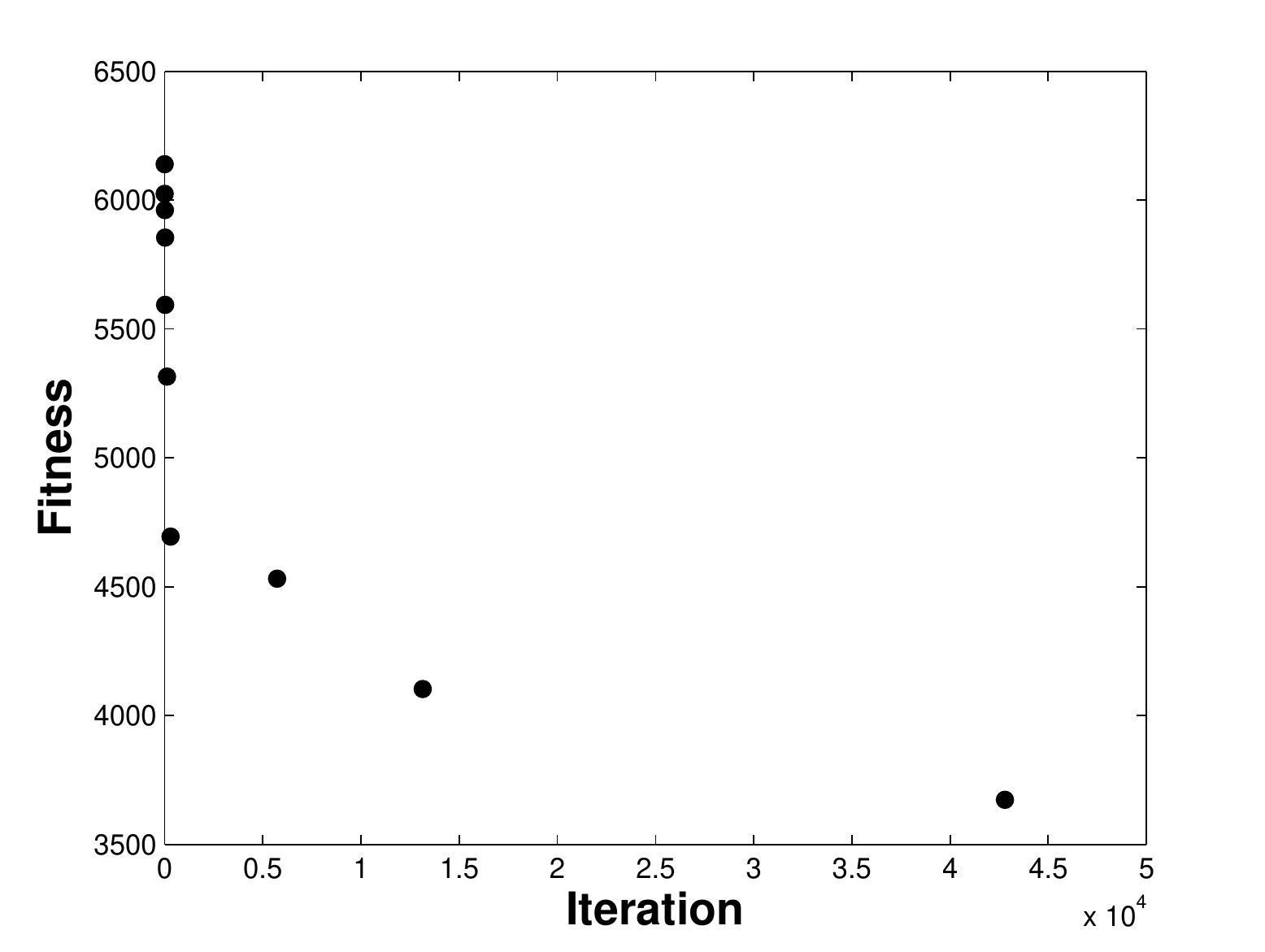}
	\caption{A purely random search continues to locate improved solutions.  Each plotted point indicates that a new global best position has been located. Logically, there is no reason that a random search would not continue to find improving solutions. Improvements are less frequent than with the \PSOC{} algorithm.}\label{fig:RANDOMUNIFORM_ContinuingImprovements}
\end{figure}

\begin{figure}[h!]
	\centering
	\includegraphics[scale=0.5]{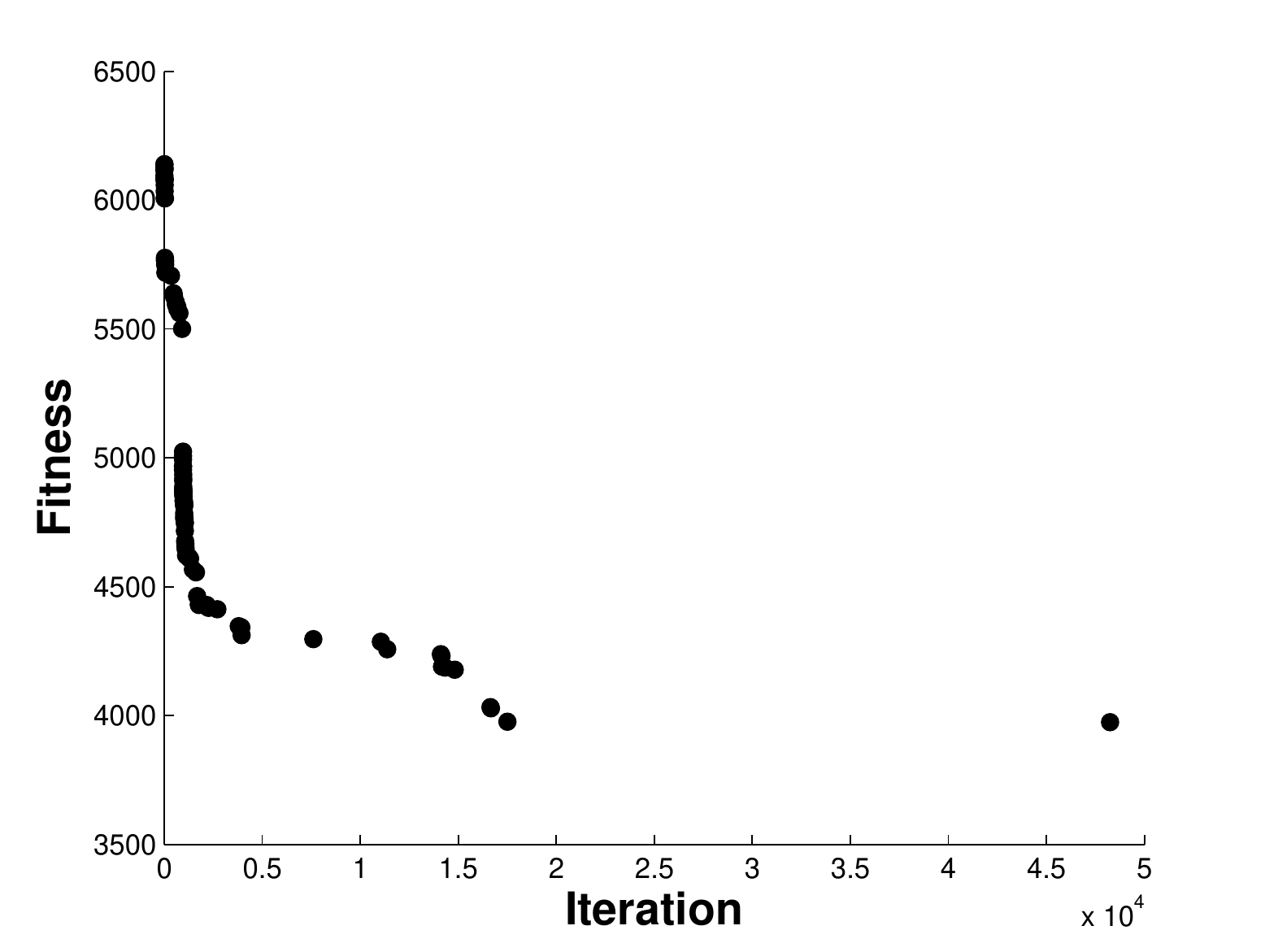}
	\caption{Performing power-law distributed random searches. Each plotted point indicates that a new global best position has been located. Best locations are only updated when a better location is found. Logically, there is no reason that a random search would not continue to find improving solutions. Improvements are less frequent than with the \PSOC{} algorithm.}\label{fig:RANDOMPOWERLAW_ContinuingImprovements}
\end{figure}

Random search would also avoid stagnation but does not leverage knowledge learnt of the objective function. 
We compared \PSOC{} with a simple uniform random search and a random power-law distribution search. 
In the former, we simply select locations within the problem space at random and evaluate them. In the latter, we maintain a set of particles which are located at the best location they have individually seen. We select new locations centered on the existing positions  with positional changes draw from a power-law distribution scaled so that all space can be reached. By selecting position changes from a power-law distribution we should ensure that the opportunity of good locations is exploited whilst it remains possible for the swarm to explore areas previously not searched. Figures \ref{fig:RANDOMUNIFORM_ContinuingImprovements} and \ref{fig:RANDOMPOWERLAW_ContinuingImprovements} show that whilst the random search do avoid stagnation and will find better sites in time, they will do so less frequently than the \PSOC{} algorithm.

\subsection{Coverage of search space}

\begin{figure}[htb]
	\centering
	\includegraphics[scale=0.5]{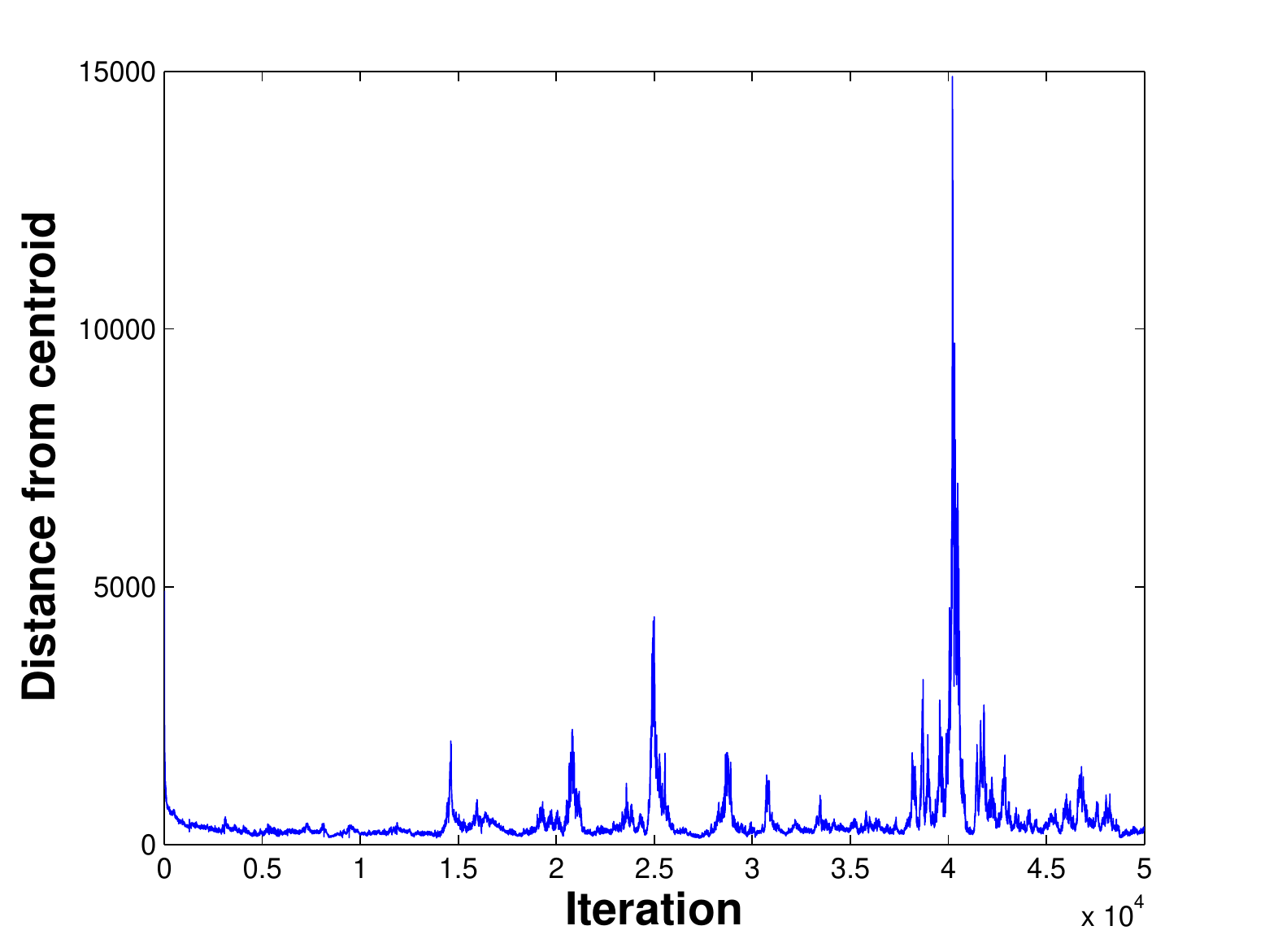}
	\caption{Mean swarm distance from the swarm's centroid. Maximum swarm size can be seen to be larger than the size of the problem's region of interest.  }\label{fig:PSOC_MSD}
\end{figure}

\begin{figure}[htb]
	\centering
	\includegraphics[scale=0.35]{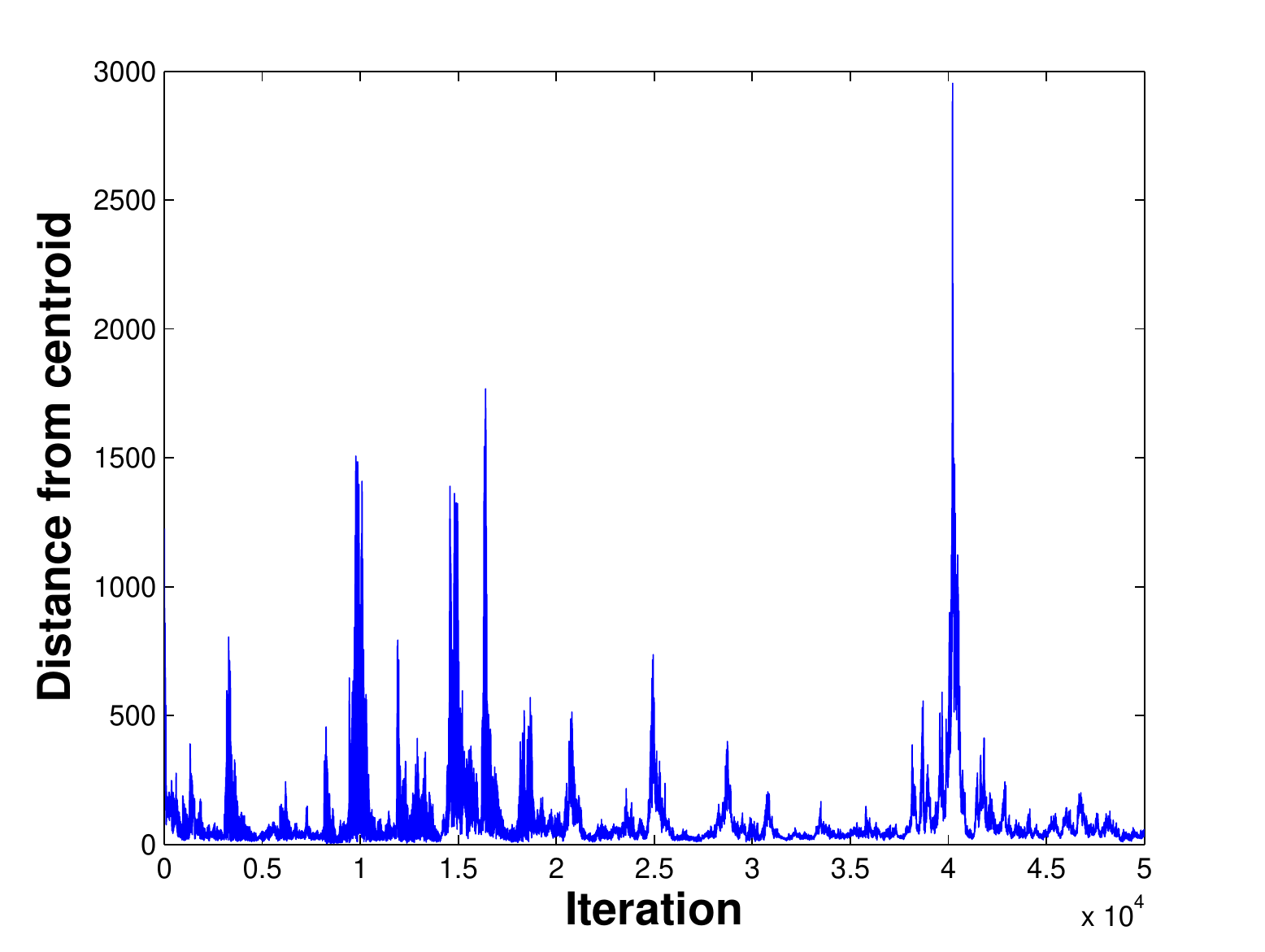}
	\includegraphics[scale=0.35]{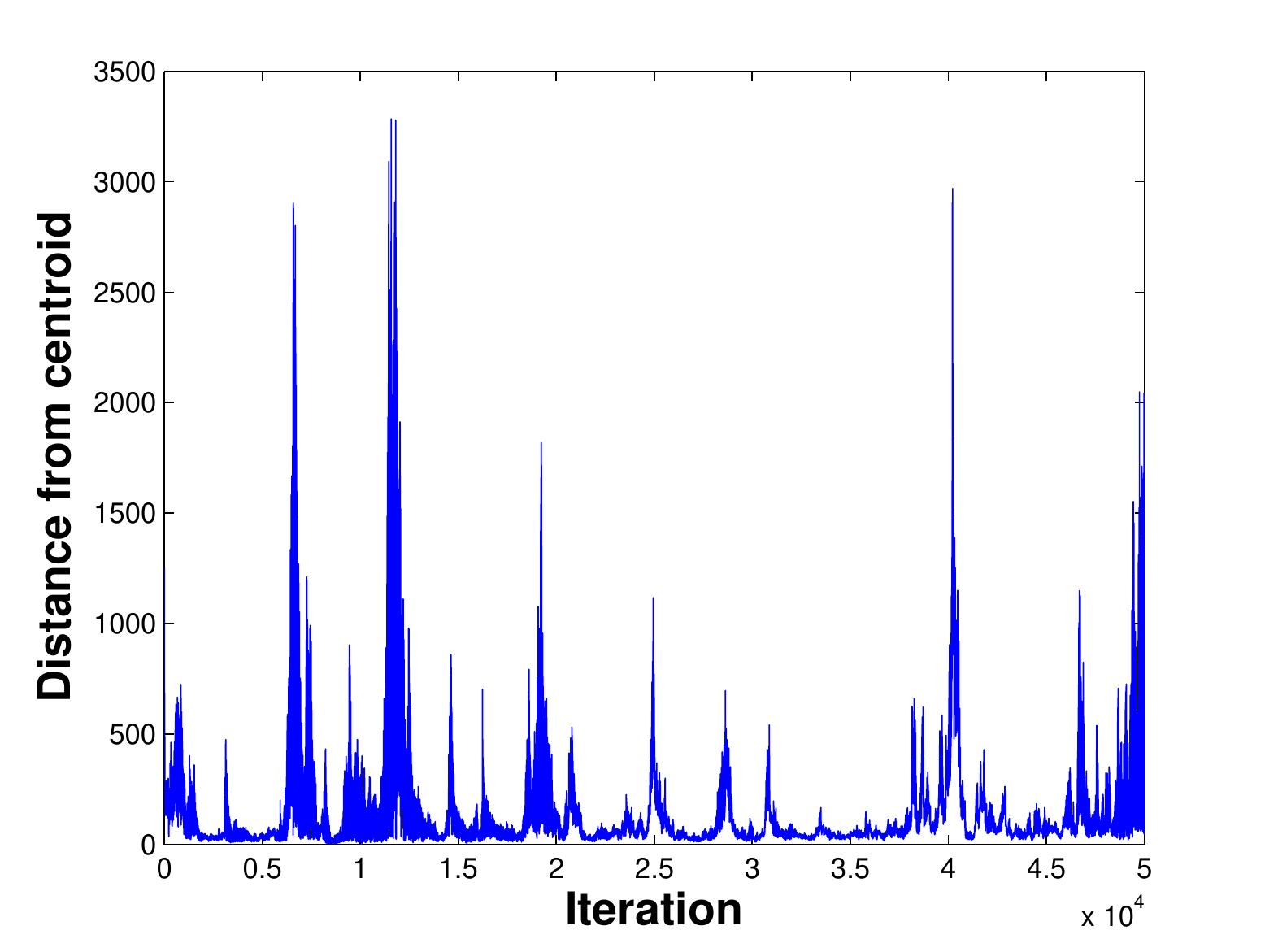}
    \includegraphics[scale=0.35]{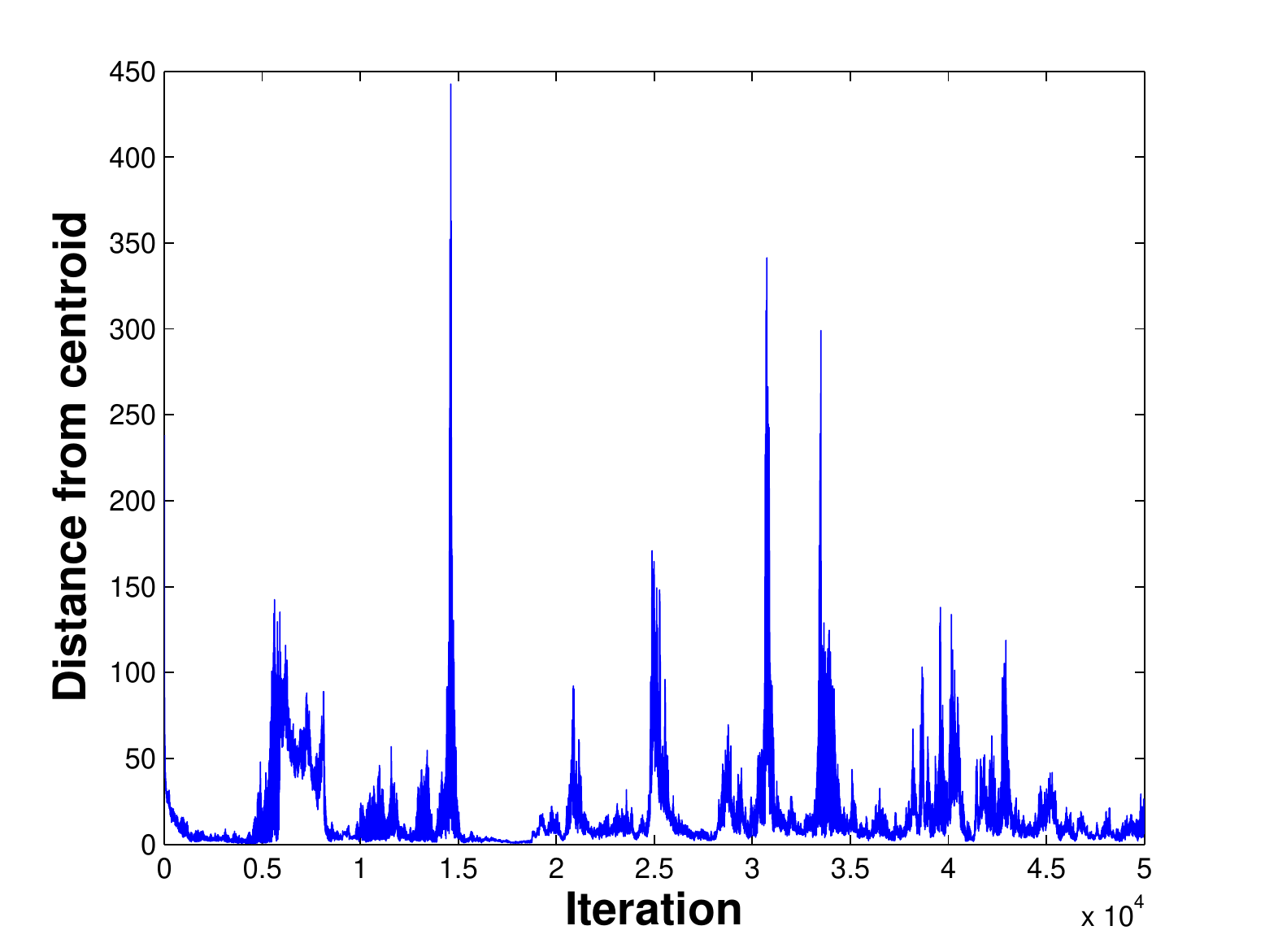}
	\includegraphics[scale=0.35]{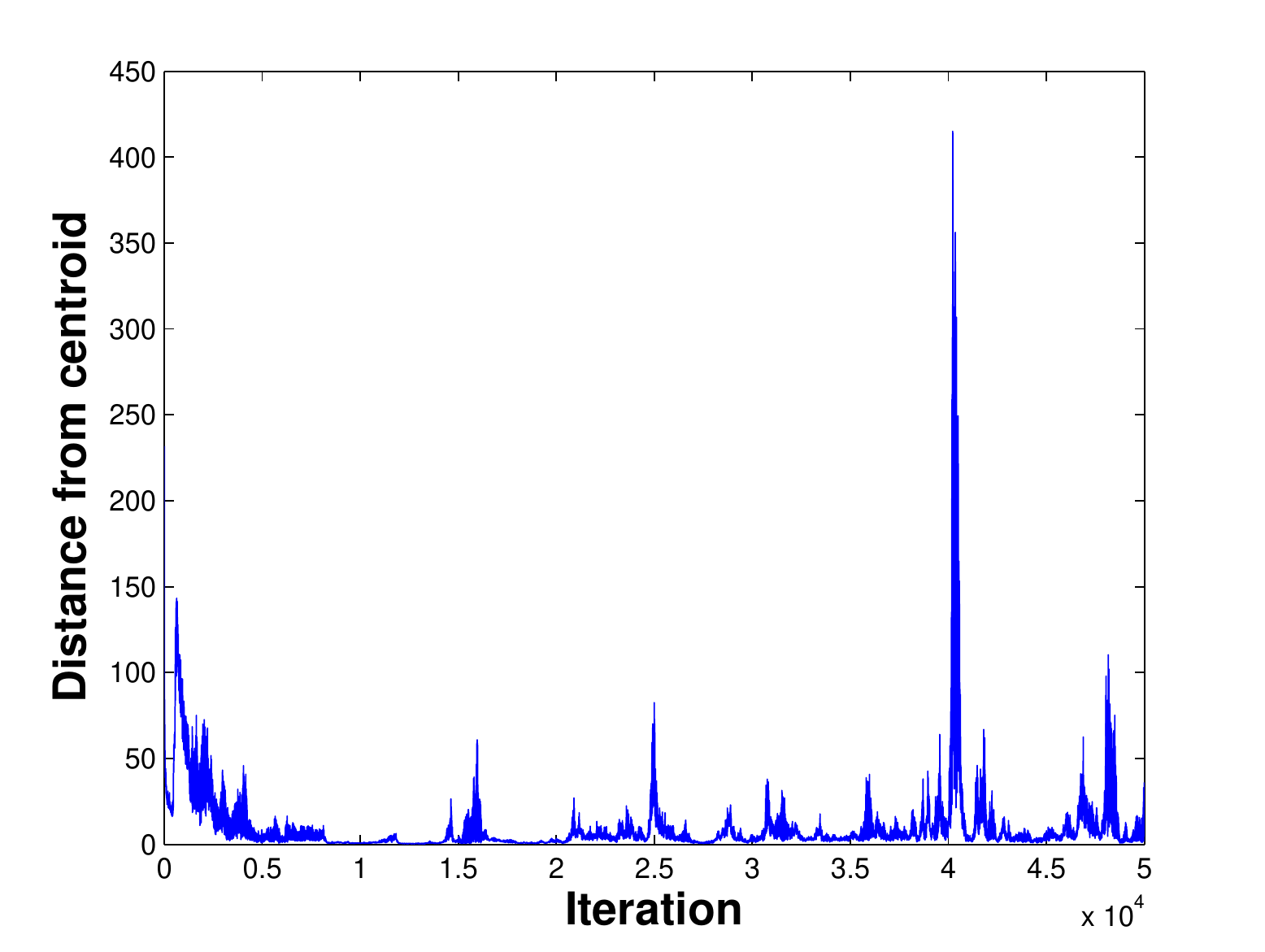}
	\caption{Different particles show different dynamics. Particle distance from swarm centroid is shown in the upper plots for two different particles. The particles are doing different things at different times. The lower plots show the dynamics of the mean swarm distance in two of the twenty dimensions. The dynamics of the swarm size is different in different times and dimensions.}\label{fig:ASYNC_BEHAVIOURS}
\end{figure}

We calculate the mean distance of the particles in the swarm from the centroid of the swarm at each iteration. As our problem space is of size 1000 in each of 20 dimensions, the maximum distance across our problem space is $\sqrt{20}\times 1000 \approx 4472$. Figure~\ref{fig:PSOC_MSD} shows the time evolution of the mean swarm distance from the swarm centroid. It is obvious that the swarm can become larger than the problem size. This is not unexpected as the limits of the search space can only be inferred by evaluating the fitness values that are strictly increasing outside the native domain of the Schwefel function.
We also see that the swarm spends larger periods of time in a compact shape than in an expanded form, which implies the 
typical picture of much exploitation with bursts of exploration. 
The dynamics of the swarm is nevertheless complex as different particles explore or exploit the space in different ways, 
similarly the balance of these behaviours is also different in different dimensions. The dimension- and particle-specific dynamics are not synchronised as shown in the plots in Fig.~\ref{fig:ASYNC_BEHAVIOURS}

\subsection{Assessment of criticality}
\begin{figure}[htb]
	\centering
	\includegraphics[width=0.5\textwidth]{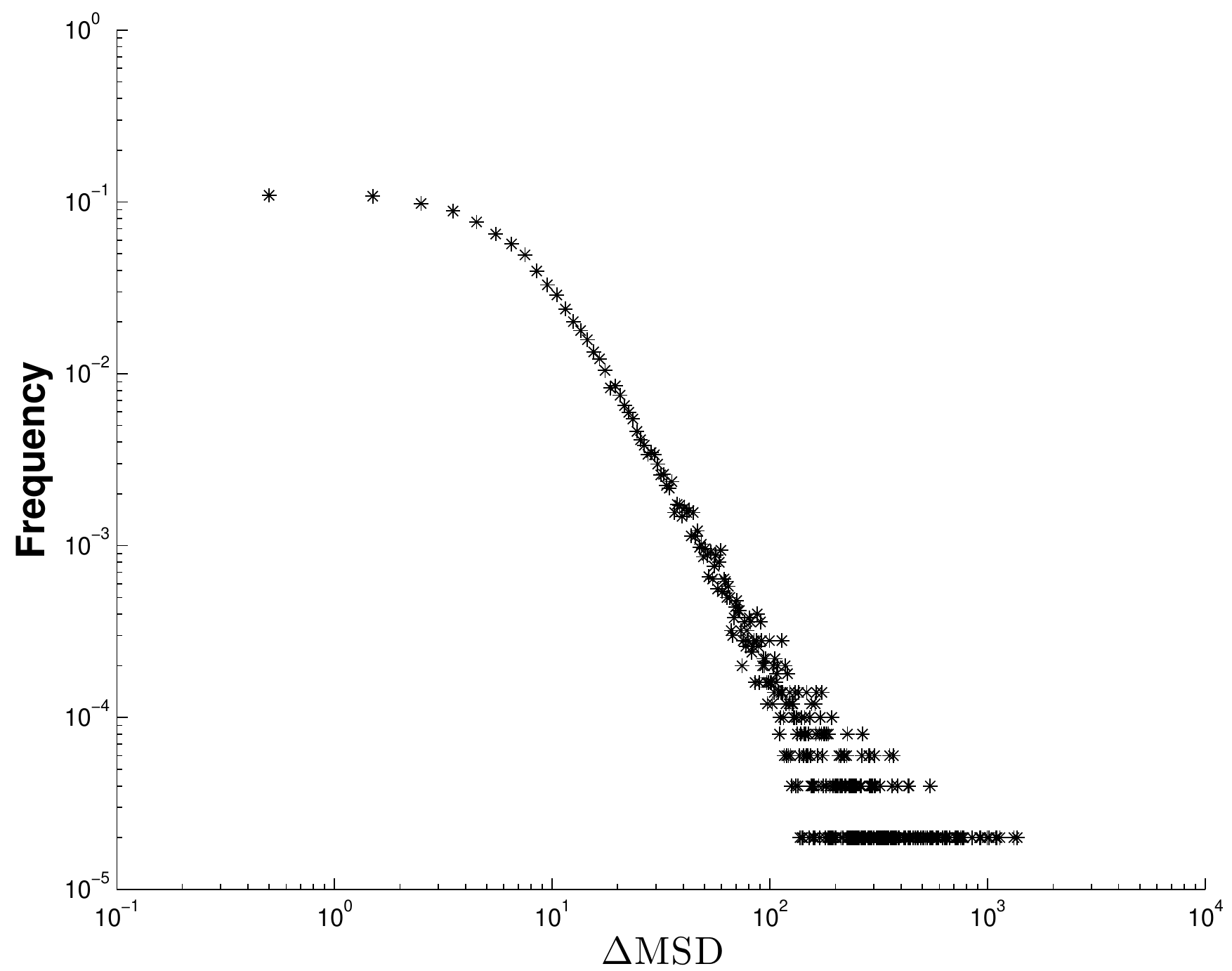}
	\caption{Large swarm dynamic shows critical behaviour. Plotting the frequency of swarm size changes shows a straight line on a log-log plot. This suggests that there may be a power-law present.}\label{fig:PSOClogloge015}
\end{figure}

Investigating the presence of power-laws requires also the study of larger PSO systems. We increased the 
number of particles in the swarm to 250 and studied the dynamics for 50000 iterations. 
All executions located the target minima to within a small error ($<$0.001).
The \PSOC{} algorithm avoids stagnation by showing a power-law distribution in the jump sizes of the swarm size. Figure~\ref{fig:PSOClogloge015} gives limited evidence for a power-law of the form
\begin{equation}\label{eq:powerlawrelationship}
y\approx 83000\, x^{-2.3}.
\end{equation}
Although it is not unusual that event distributions show a deviation from a power-law outside a lower and an upper cut-off,
it is often required that the distribution behaves linearly in the log-log plot for at least two decades
which is not reached here. It is possible that we need to increase the particle numbers or the size of the 
region of interest to allow any power-law to become  apparent over larger scalings.



\begin{figure}[h]
\centering
\includegraphics[width=0.5\textwidth]{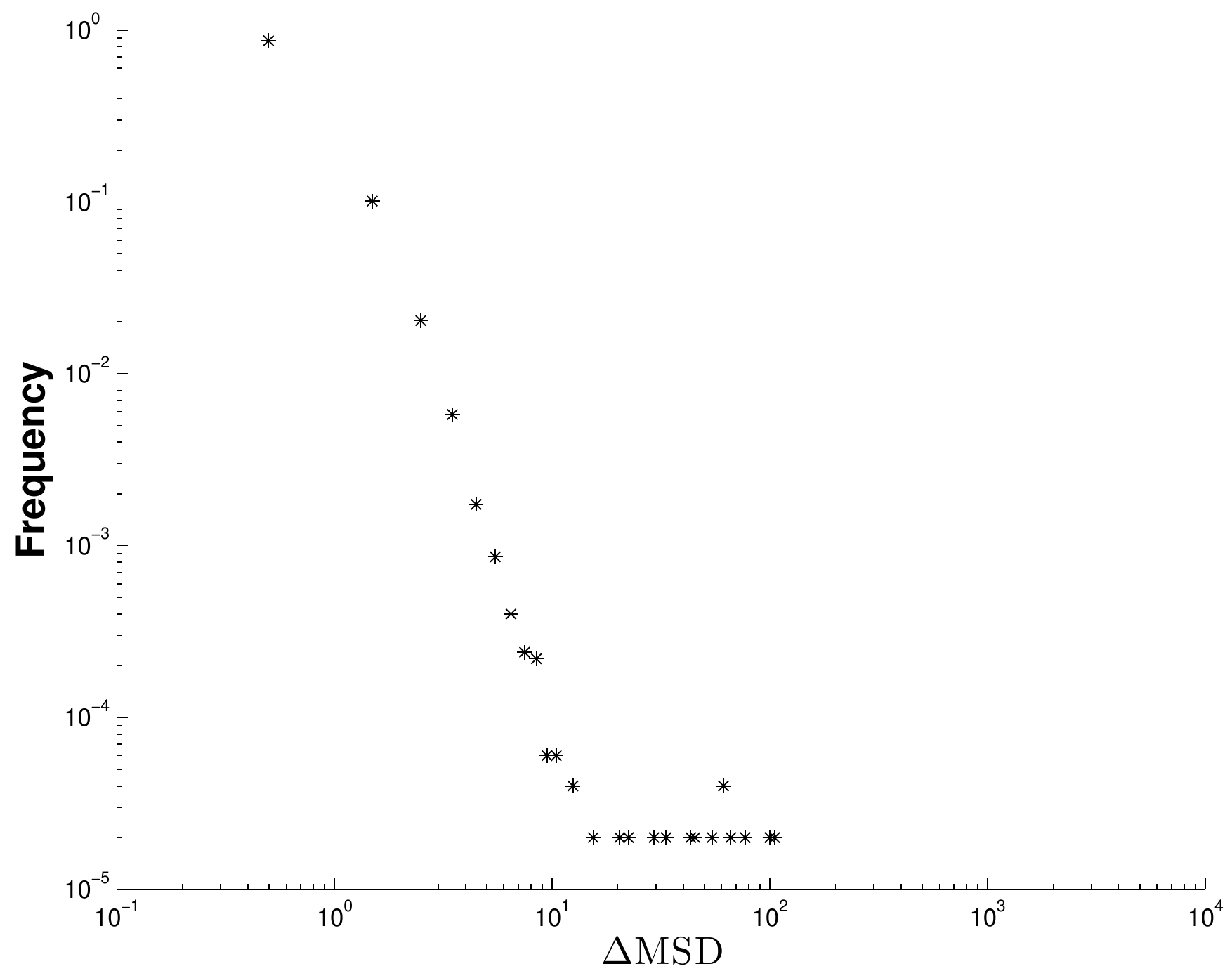}

\includegraphics[width=0.5\textwidth]{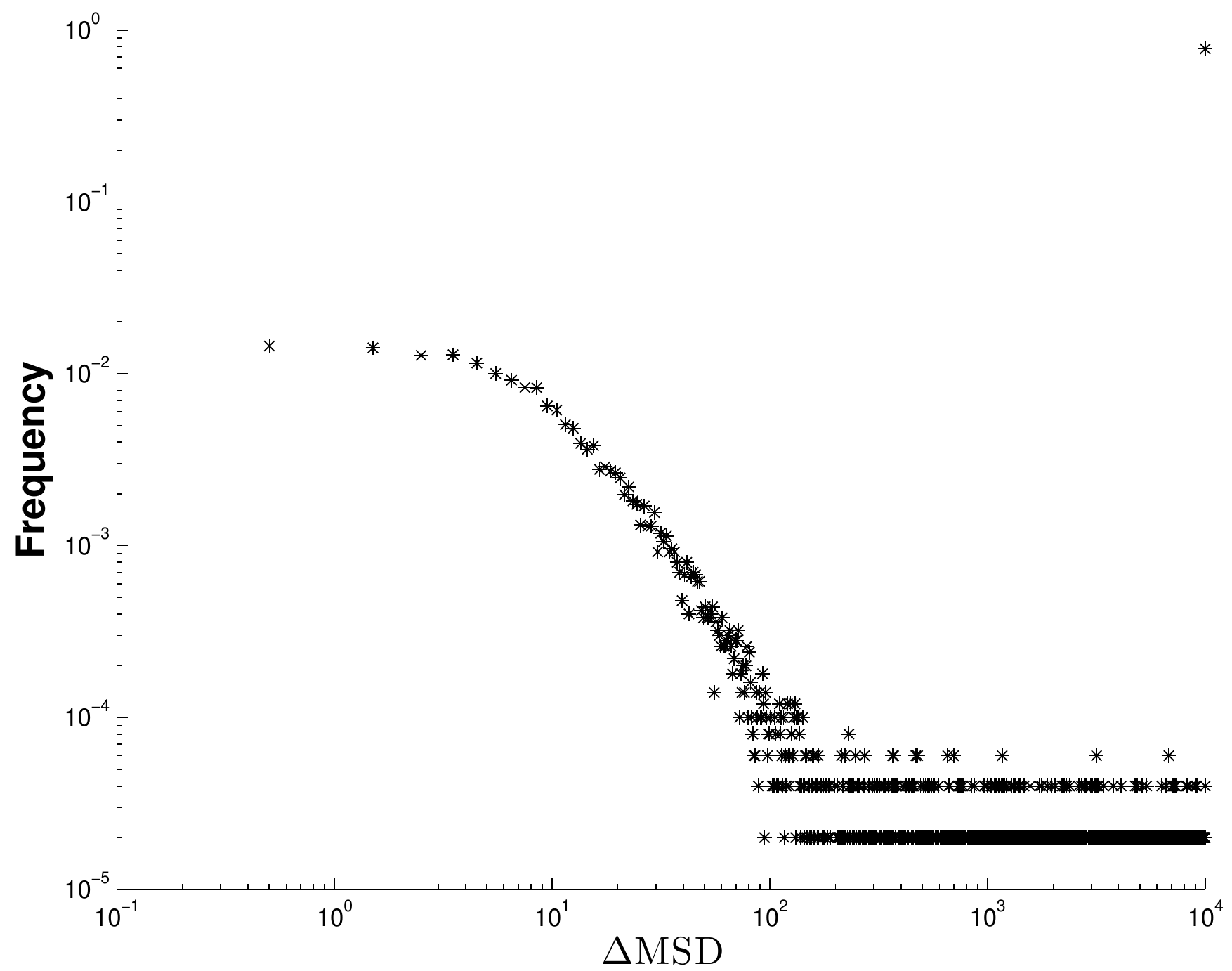}

\includegraphics[width=0.5\textwidth]{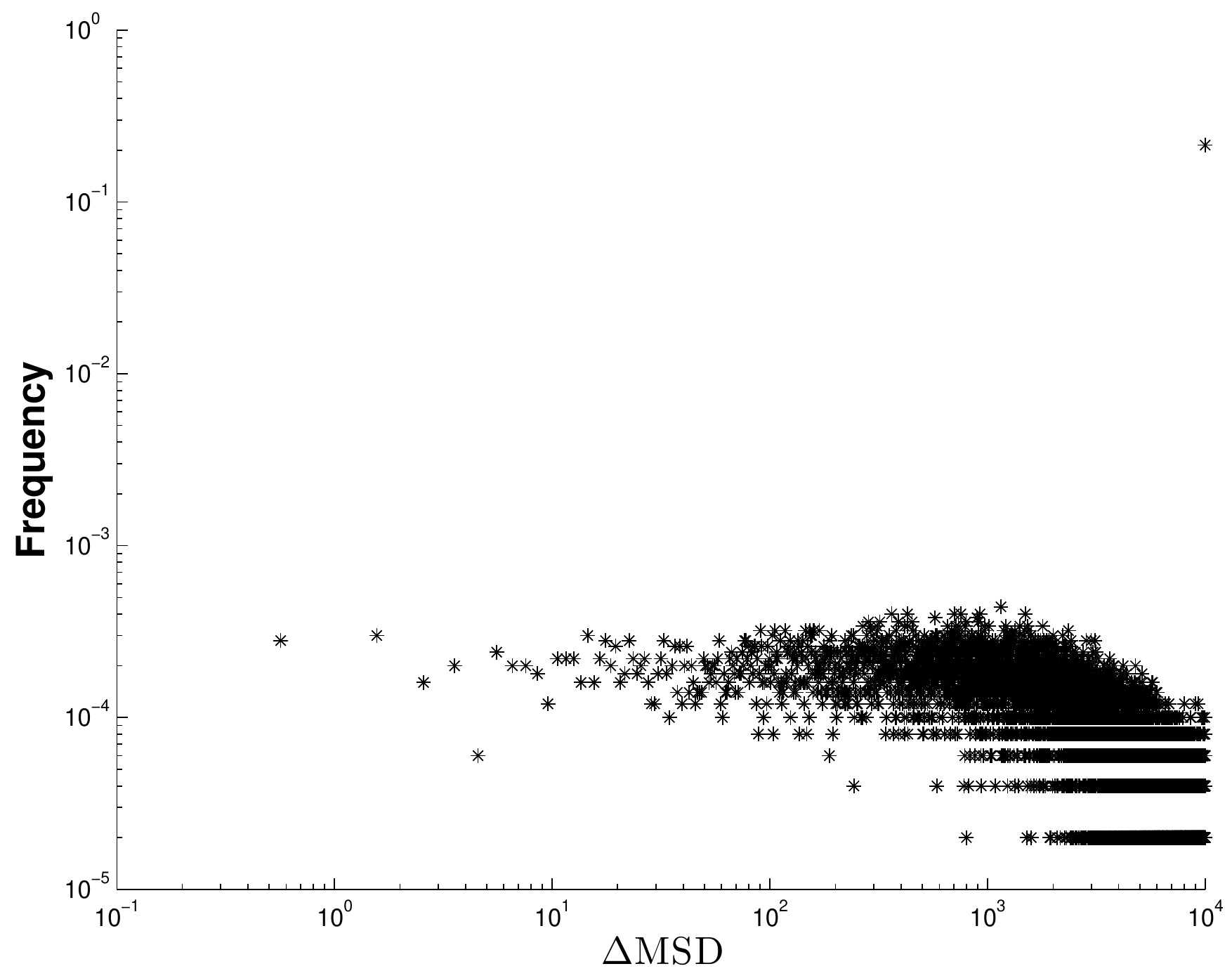}
\caption{Modifying the $\varepsilon$ parameter changes the behaviour of the algorithm. The top plot ($\varepsilon$=0.5) shows that the swarm makes no large changes in size such that exploration of the problem space remains limited. The middle plot ($\varepsilon$=0.075) shows the mean swarm diameter 
jump sizes. 
Finally, for comparison, the bottom plot shows the equivalent plot for the standard PSO algorithm (with $\omega$=0.7, $\alpha_1$=2, and $\alpha_2$= 2. Note that the
last bin of the histogram shows all swarm sizes above the upper bound. For the top plot this number is zero.}\label{fig:PSOCloglog}
\end{figure}

If the straight line represents a true power-law relationship then we should suspect that it is possible to tune the system via a parameter variation from subcritical, through critical, to supercritical behaviours. 
The starting values of the parameters $\omega$, $\alpha_{1}$, and $\alpha_{2}$ appear to make little difference 
to the behaviour of the algorithm. 
This is to be expected as the algorithm is itself modifying these parameters. 
Instead we look at the $\varepsilon$ parameter. Large values ($\varepsilon = 0.5$) appear to result in a subcritical swarm with a shortage of large swarm movements, whilst a small value ($\varepsilon = 0.075$) results in an excess of large moves making the swarm appear supercritical.  Finally the equivalent plot for the standard PSO algorithm. With the parameters used the PSO swarm makes many large moves, favouring exploration over exploitation. This confirms the earlier observation that parameter's used for out PSO comparison were poor. Choosing different values may have improved the PSO's performance. Fig.~\ref{fig:PSOCloglog} shows these variants.

\section{Discussion}\label{sect:disc}
The \PSOC{} algorithm can be tuned via its $\varepsilon$ parameter to operate in a self-organised critical manner. With the correct setting, the swarm is able to explore the full region of interest of the test function. The extension of the swarm varies with a power-law distribution. As the standard PSO element of the algorithm draws the swarm toward the current best locations these size changes assure that areas of the problem space near known good values are well exploited. The long tail of the power-law distribution will ensure that exploration of the full problem space occurs. These two competing mechanism stop the algorithm from stagnating. 

Our studies of the \PSOC{} algorithm applied to the Schwefel test function suggest that it is  insensitive to the initial settings of the $\omega$, $\alpha_1$, and $\alpha_2$. This is expected since the algorithm is continually modifying the parameter values throughout its execution. The range over which this insensitivity extends shall be explored in future work. The important parameter to set is $\varepsilon$. Varying this parameter tunes the behaviour of the swarm from subcritical to supercritical behaviours. For other problems where the optimum values are unknown we believe that detection of critical behaviour should be easier to achieve than correctly setting the $\omega$, $\alpha_1$, and $\alpha_2$ parameters. Further investigation of the relationship between $\varepsilon$ and the power-law exponent should be done.

We have reviewed the capabilities of the algorithm largely in the terms of total number of particle iterations. The algorithm, because it calculates the change in swarm size, $\Delta S$, and a sigmoid squashed function from this value, has a higher computational overhead than standard PSO. It is not clear the sigmoid squashing function is needed. Removing it or replacing it with a simplified value lookup would reduce this overhead. The standard PSO use of random stochastic variables in the velocity update was included in the \PSOC{} algorithm. It is possible that the critical dynamics of the \PSOC{} algorithm may itself remove the need to include these.

The \PSOC{} algorithm, as explored, requires there to be no boundary to the problem space or limits to particle velocities. We noted that if these freedoms were applied to our implementation of standard PSO improvements were also seen (although not as great as the \PSOC{} improvements) in terms of continued exploration of the problem space.

The \PSOC{} algorithm, with its $\varepsilon$ parameter set to ensure self-organised critical behaviour showed an automatic balancing of exploitation and exploration of the problem space. Stagnation is avoided resulting in improving results the longer the algorithm is run. The mechanism that gives rise to this links the interaction of the swarm with its problem environment to modification of the algorithm's own parameters. In this way the algorithm is tuning itself to the problem space.  This ability to self-tune the parameters the \PSOC{} algorithm should then be demonstrated across a much wider range of test and real world functions.

In the earlier work \citep{cordero2012} the particle velocities had an upper bound applied (as is common with many PSO variants). This would impact the size of changes to the swarm size between iteration. Some critical behaviours were seen in the dynamics on individual particles. The inclusion of an upper velocity will limit the range over which such behaviours may be seen. Our algorithm allows our particles greater freedom by removing this constraint.




\section{Conclusions}
We have presented an algorithm for metaheuristic optimisation that employs a critical particle 
swarm (\PSOC). It achieves a balance between exploration of the problem space and exploitation of the objective. 
The algorithm does not use any other information about the objective function than the standard version of PSO.
Instead \PSOC{} self-adapts its parameters based on holistic information about the swarm.
We have used the average velocity norm, but other properties of the swarm such as local density, 
swarm extension or gradient of the objective function within the swarm can be used as well. 
The PSO parameters are obtained from the swarm properties by a simple control algorithm. It would clearly 
be interesting to study more advanced control schemes.
Furthermore, we have used the information homogeneously across the swarm, 
i.e.~although the parameters are temporally variable, they are identical for all particles at any 
given moment of time. Heterogeneous swarms~\citep{Sayama2009SwarmChemistry} are known 
to produce a much richer manyfold of behaviours which appears to be promising
not only for modelling of biophysical systems~\citep{Erskine2013ECAL} but potentially also for application
in optimisation, see~\citep{engelbrecht2010heterogeneous}. 
In particular it might be interesting to move towards an ecological optimisation algorithm
that can be developed from a combination of \PSOC{} and the 
TRIBES approach \citep{clerc2003tribes}. 




\begin{acknowledgements}
The authors acknowledge the contribution of Carlos Garcia Cordero submitted an excellent study of an earlier version of the algorithm~\citep{cordero2012}. This work was supported by the Engineering and Physical Sciences Research Council [grant number EP/K503034/1].\end{acknowledgements}

\bibliographystyle{spbasic}
\bibliography{PSOC_S2}

\end{document}